\documentclass[10pt,twocolumn,letterpaper]{article}

\usepackage[pagenumbers]{cvpr} 

\usepackage{graphicx}
\usepackage{tabularx}
\usepackage{makecell}
\usepackage{rotating}
\usepackage{array}
\usepackage{amsmath}
\usepackage{amssymb}
\usepackage{booktabs}
\usepackage{pgffor}
\usepackage{adjustbox}
\usepackage{multirow}
\usepackage{bbding}
\usepackage[ruled]{algorithm2e}

\usepackage[pagebackref,breaklinks,colorlinks]{hyperref}

\usepackage[capitalize]{cleveref}
\crefname{section}{Sec.}{Secs.}
\Crefname{section}{Section}{Sections}
\Crefname{table}{Table}{Tables}
\crefname{table}{Tab.}{Tabs.}

\def\sammed{{SAM\(^{\text{Med}}\)~}}
\def\assist{{SAM\(^{\text{assist}}\)~}}
\def\auto{{SAM\(^{\text{auto}}\)~}}

\def\cvprPaperID{*****} 
\def\confName{CVPR}
\def\confYear{2022}

\begin{document}

\title{\sammed: A medical image annotation framework based on large vision model}

\author{Chenglong Wang \textsuperscript{1} \qquad Dexuan Li \textsuperscript{1} \qquad Sucheng Wang \textsuperscript{1} \qquad Chengxiu Zhang \textsuperscript{1} \\ Yida Wang \textsuperscript{1}  \qquad Yun Liu \textsuperscript{2} \qquad Guang Yang \textsuperscript{1} \\
\begin{tabular}{cc}
\textsuperscript{1} Shanghai Key Laboratory of Magnetic Resonance, & \multirow{2}{*}{\textsuperscript{2} NVIDIA \qquad} \\
East China Normal University &                  
\end{tabular}\\
{\tt\small \{clwang, gyang\}@phy.ecnu.edu.cn  \quad yunl@nvidia.com
}
}

\maketitle

\begin{abstract}
Recently, large vision model, Segment Anything Model (SAM), has revolutionized the computer vision field, especially for image segmentation. SAM presented a new promptable segmentation paradigm that exhibit its remarkable zero-shot generalization ability. An extensive researches have explore the potential and limits of SAM in various downstream tasks. In this study, we presents \sammed
, an enhanced framework for medical image annotation that leverages the capabilities of SAM.

\sammed framework consisted of two submodules, namely \assist and
\auto. The \assist demonstrates the generalization ability of SAM to the downstream medical segmentation task using the prompt-learning approach. Results show a significant improvement in segmentation accuracy with only approximately 5 input points. The \auto model aims to further accelerate the annotation process by automatically generating input prompts. The proposed SAP-Net model achieves superior segmentation performance with only five annotated slices, achieving comparable segmentation performances with \assist. Overall, \sammed demonstrates promising results in medical image annotation. These findings highlight the potential of leveraging large-scale vision models in medical image annotation tasks.
\end{abstract}

\section{Introduction}
\label{sec:intro}

In recent years, Large Language Models (LLMs) have witnessed remarkable success and garnered considerable attention from both academic and industrial domains. Notably, ChatGPT stands out as one of the most prominent model of LLMs, showing its remarkable ability to comprehend and generate human-like language. It has fully demonstrated the potential and extraordinary scalability of large models. For the first time, it showcased the remarkable potential of the General AI to comprehend and process human language with an unprecedented level of proficiency. This groundbreaking achievement not only revolutionized natural language processing but also influenced computer vision field. 

More recently, Kirillov \etal proposed ``Segment-Anything Model'' (SAM) \cite{kirillov2023segany}, a large vision model constructed on the largest segmentation dataset to date, with over 1 billion masks on 11M licensed natural images. Notably, SAM exhibited exceptional zero-shot transfer capabilities, enabling its adeptness in generalizing to previously unseen data. By defining a novel promptable segmentation task, SAM effectively segments images based on specific segmentation prompts, such as points and bounding boxes that specify the target objects. This approach provides SAM with a unique advantage in accommodating varied input cues and facilitating precise and various object segmentation, contributing to its remarkable performance in challenging scenarios. The incorporation of prompt-based segmentation enriches its repertoire of capabilities, making it an invaluable asset in diverse image analysis applications.

\subsection{Medical Annotations}
Medical image annotation is a critical task in the field of healthcare and medical research. Accurate and efficient annotation of medical images is essential for various down-stream applications such as disease diagnosis, tumor detection. Traditional annotation tools such as ITK-SNAP \cite{py06nimg}, 3D Slicer \cite{fedorov20123dslicer} has been commonplace in the medical imaging community. These tools focus on providing stable annotation platform for the users. With the development of deep-learning techniques in medical segmentation, several advanced image annotation tools have been developed using neural network as segmentation backbones, such as RITM \cite{sofiiuk2022ritm}, SimpleClick \cite{liu2022simpleclick}, and MIDeepSeg \cite{LUO2021102102}. 

\subsection{About Segment-Anything Model}
The Segment Anything Model (SAM) is comprised of three core components: \textit{image encoder}, \textit{mask decoder} and \textit{prompt encoder}. The image encoder uses a heavyweight vision transformer-based architecture to effectively extract image features. In this regard, SAM provides three pre-trained image encoders, each tailored to different scales, denoted as ViT-B, ViT-L, and ViT-H, incorporating 91M, 308M, and 636M parameters, respectively. The pre-trained image encoder plays a pivotal role in embedding high-resolution images (with dimensions of $1024\times1024$) into low-resolution image embedding (with dimensions of $64\times64$). 

The mask decoder, on the other hand, adopts a lightweight transformer network, enabling the generation of precise image segmentation results based on the extracted image features and the given prompt information. The mask decoder incorporates two transformer layers, a dynamic mask prediction head, and an Intersection-over-Union (IoU) score regression head.

To facilitate a comprehensive representation of users' input prompt details, prompt encoders are strategically designed. SAM supports four different prompt types: points, boxes, texts, and masks, with each type employing specific encoding methods to ensure optimal representation. For instance, points and boxes are encoded by Fourier positional encoding \cite{tancik2020fourier} combined with two learnable tokens for foreground and background. Mask prompt preserves the spatial resolution identical to that of the input image, and encoded by convolution operation. The text prompt is encoded by the pre-trained text-encoder in CLIP \cite{radford2021clip}.

\subsection{SAM in medical imaging} \label{sec:sam_med}
SAM has revolutionized the field of image segmentation field, facilitating zero-shot processing and enabling further development for diverse downstream tasks. Within the domain of medical image processing, SAM has emerged as a focal point of research, evidenced by a substantial number of preprint papers.

Among this corpus of literature, several studies have exclusively focused on leveraging SAM for segmentation tasks in specific medical domains. These include pathology image segmentation \cite{deng2023segment}, liver tumors segmentation in CT scans \cite{hu2023sam}, abdominal CT image segmentation \cite{wald2023sam}, brain tumor segmentation in MRI \cite{putz2023segment}, polyp detection \cite{zhou2023polyp, li2023polyp}, and ophthalmology image segmentation \cite{qiu2023learnable}. Concurrently, several works have conducted more comprehensive evaluations, comparing SAM-based segmentation across multiple organ sites and various medical imaging modalities \cite{mazurowski2023segment,cheng2023sam}.

Remarkably, Mazurowski \etal have conducted a comprehensive evaluation experiments to date. They found that drawing box prompt on each object yielded the most optimal segmentation performance in medical imaging. This findings are in strong agreement with other studies \cite{mazurowski2023segment}. Beyond investigating segmentation performance, Ma \etal and Li \etal fine-tuned the SAM to address specific medical segmentation application \cite{ma2023segment,li2023polyp}. Their work demonstrated the significant efficacy of fine-tuning the SAM on medical images, resulting in enhanced segmentation accuracy for medical imaging purposes.

Furthermore, SAM's influence extends to diverse applications, as highlighted by a notable study that successfully integrated SAM into 3D Slicer. This integration has substantially enhanced the utility and accessibility of SAM within the medical imaging community.

\subsection{Contributions}

\begin{enumerate}
    \item In this study, we propose \sammed framework consisted of two submodules, namely \assist and \auto, to harness the remarkable generalization capabilities of large vision models, SAM, to accelerate the medical image annotation process. By leveraging the power of SAM model, we significantly reduce the complexity and time-consuming for annotating medical images, thereby enhancing the efficiency of the overall annotation workflow.

    \item The proposed \assist model employs the prompt-learning technique to effectively adapt the down-stream medical segmentation task to the SAM. The utilization of prompt-learning enables SAM to achieve remarkable performance improvements with minimal additional computational resources, making it an efficient and effective solution for medical image annotation tasks.

    \item The proposed \auto model is proposed to further boost the annotation process. By introducing a prompt generator module, it can effectively reduce the manual interaction process. We investigate two different approaches for the automatic prompt generation. We also propose a spatial-aware prototypical network, namely SAP-Net, for accurate prompt generation. By harnessing the power of few-shot learning, the SAP-Net is capable of automatically generating prompts required by SAM.
\end{enumerate}

\section{Related works}\label{sec:related}

Since the medical image annotation is widely recognized as a formidable task that poses significant challenges for human experts in the field. Consequently, a mount of research efforts have been dedicated to tackling this problem with the aim of finding effective solutions.

One approach that has been explored is the utilization of minimal user interaction, specifically through extreme point clicks, to alleviate the burden on human experts \cite{LUO2021102102, make3020026}. Roth \etal employed extreme points as attention mechanisms to enhance the segmentation results of an initial coarse segmentation generated by random walker algorithm. Similarly, Luo \etal leveraged extreme points to generate a distance map as an external cue for a neural network, and refine the initial segmentation through additional user clicks \cite{LUO2021102102}.

Another solution is to use minimal annotated data to accomplish comparable segmentation accuracy compared to fully-supervised segmentation. Roy \etal incorporated a few-shot learning methodology to train a volumetric segmentation model using a limited number of annotated support examples, demonstrating a zero-shot transfer capability \cite{GUHAROY2020101587}. Cai et al. proposed a semi-supervised technique that incorporates orthogonal annotated slices as a consistency term, addressing both weak annotation and semi-supervised challenges in medical image segmentation \cite{cai2023orthogonal}.

Other annotation frameworks, such as MONAI Label \cite{monailabel}, and AbdomenAtlas \cite{qu2023annotating}, leverage the capabilities of pre-trained medical segmentation model to facilitate an efficient human-in-the-loop annotation procedure. The efficacy of these fully-supervised approaches heavily depend on the pre-trained models, which consequently limits their ability for zero-shot segmentation. In cases where annotating an unseen class is required, retraining or fine-tuning the underlying model becomes necessary.

As introduced in \Cref{sec:sam_med}, an increasing number of studies have incorporated SAM into their specific medical image segmentation tasks. Among these works, MedSAM\cite{ma2023segment}, MSA\cite{wu2023msa} and SAM-Med2D\cite{cheng2023sammed2d} fine-tuned the SAM to adapt it for various medical segmentation applications. Notably, a significant improvement was achieved by fine-tuning SAM's mask decoder or image encoder models. However, similar to the aforementioned works, fine-tuning SAM for downstream tasks compromises the generalization ability of large model, and it demands substantgial training resources. For instance, the SAM-Med2D model collected 19.7M annotated medical masks for its fine-tuning, while MedSAM collected approximately 1.1M masks for this purpose.  In essence, these kind of works have transferred SAM into a fully-supervised medical segmentation tasks, but with additional manual prompts.

\begin{figure*}[t!]
\begin{center}
   \includegraphics[width=\linewidth]{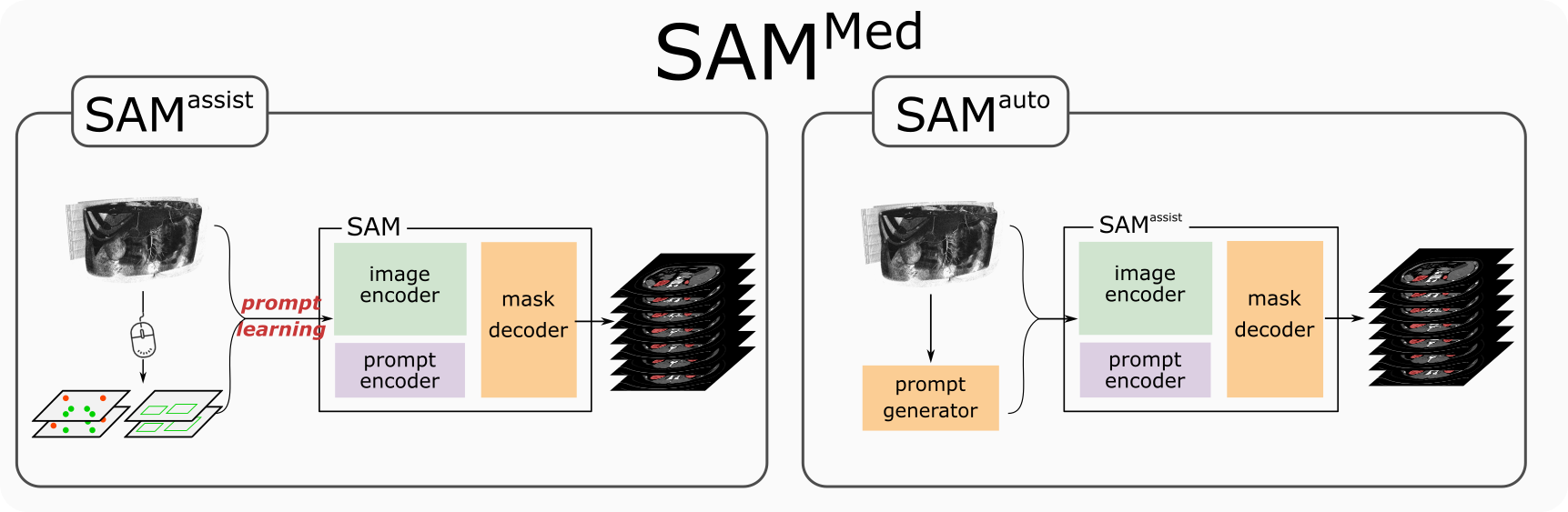}
\end{center}
   \caption{Overview of the proposed \sammed framework.
   }
\label{fig:sam_3d}
\end{figure*}

\section{Method}\label{sec:method}

In this section, we first introduce the problem setup, then detail the architecture design of our \sammed annotation tool and the training strategy, and finally describe the evaluation strategy. \sammed mainly consists of two modules: assist and auto modules, namely $\mathrm{SAM^{assist}}$ and $\mathrm{SAM^{auto}}$. $\mathrm{SAM^{assist}}$ aims to assist users to annotate medical images more efficiently. $\mathrm{SAM^{auto}}$ is designed to automatically generate the annotations without users' manual operations.

\subsection{Problem setting}\label{sec:problem}
\sammed is designed as an annotation tool for medical images. First, \sammed need initial annotations $\mathcal{D}_{Train} = \{(I_T^i, L_T^i)\}_{i=1}^N$ to train the models. Initial manually annotated $\mathcal{D}_{Train}$ comprised $N$ pairs of original image slice $I_T$ and its corresponding binary label $L_T$. The objective of $\mathrm{SAM^{assist}}$ model is to learn a model $\mathcal{F}_{assist}(\cdot)$ from $\mathcal{D}_{Train}$ so that given a new query image slice $I_Q$ and its corresponding manual input prompts $\mathbf{P}_Q$, the binary segmentation $\Tilde{L}_Q$ is inferred. The objective of $\mathrm{SAM^{auto}}$ is to learn a model $\mathcal{F}_{auto}(\cdot)$ from $\mathcal{D}_{Train}$ so that given only a new query image slice $I_Q$, the binary segmentation $\Tilde{L}_Q$ is inferred.

\subsection{Architectures}\label{sec:archi}
As previously mentioned, the proposed \sammed consists of two modules: $\mathrm{SAM^{assist}}$ and $\mathrm{SAM^{auto}}$. These two modules cooperate to maximize the efficiency of medical annotation procedure.

\subsubsection{$\text{SAM}^{\text{assist}}$} 
As many previous works \cite{mazurowski2023segment,cheng2023sam} have investigated that the original SAM model has limited performance on various medical image segmentation tasks. The segmentation performances heavily depend on the users' input prompts. Despite the impressive zero-shot performance showed by SAM in numerous vision tasks, its effectiveness is still impacted by the domain gap between natural images and medical images. 

To harness the remarkable generalization capabilities of large vision model, we incorporated \textit{prompt learning} approach to redefine the down-stream task for the large vision model. Compared to the conventional fine-tuning based approach \cite{ma2023segment,li2023polyp} which focuses on adapting the original model to fit the down-stream tasks, the prompt-learning based approach aims to tune the down-stream tasks to fit the original large model. The key advantage of prompt learning is its ability to achieve effective results with significantly reduced training data demand and training time-consuming compared to fine-tuning based approaches.

\begin{figure}[t]
\begin{center}
   \includegraphics[width=\linewidth]{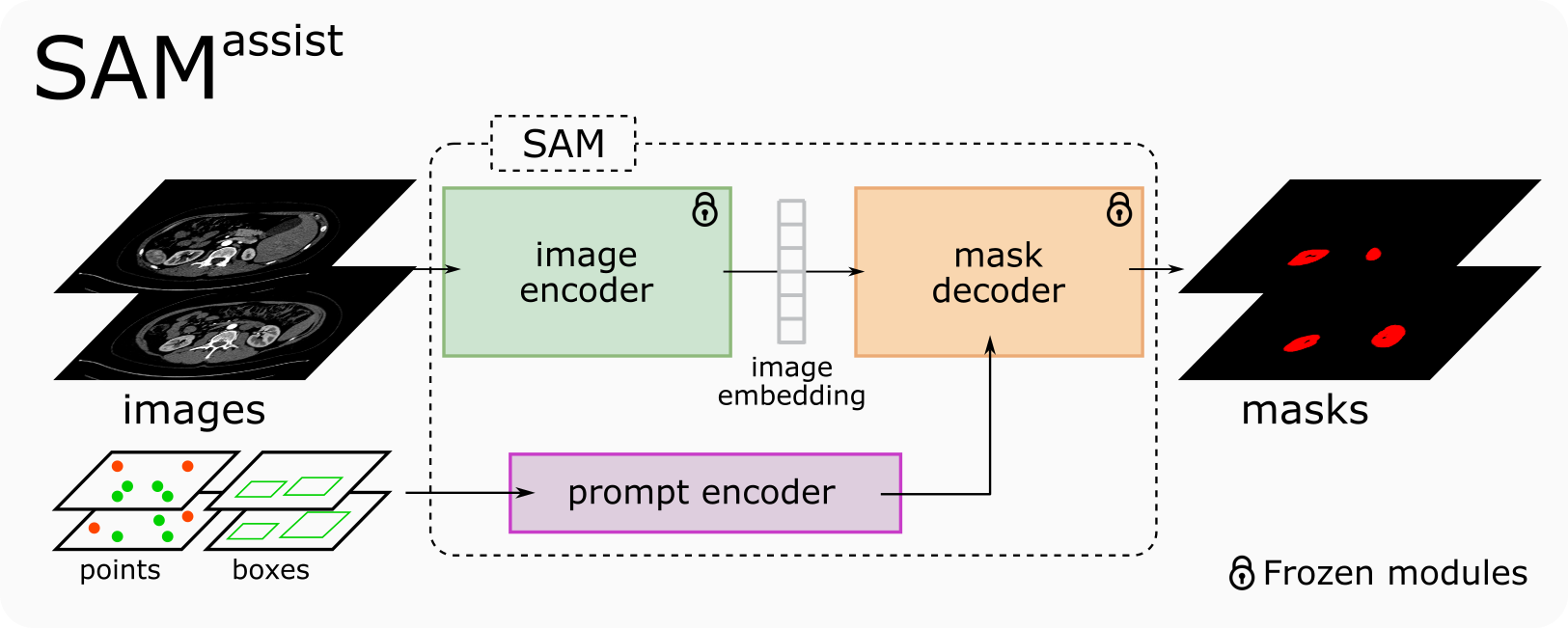}
\end{center}
   \caption{Overview of the proposed \assist model. The three modules, namely ``image encoder'', ``mask decoder'' and ``prompt encoder'', employed in this study are derived from the original pre-trained SAM model. Notice that only ``prompt encoder'' module is trainable, while ``image encoder'' and ``mask decoder'' remain frozen throughout the training phase. 
   }
\label{fig:sam_assist}
\end{figure}

The overview of \assist is illustrated at~\cref{fig:sam_assist}. The training dataset $\mathcal{D}_{Train} = \{(I_T^i, L_T^i)\}_{i=1}^N$ consists of $N$ pairs of annotated image slice $I_T$ and corresponding annotation $L_T$, $I_T^i \in \mathbf{I}_T$, $L_T^i \in \mathbf{L}_T$. The objective of \assist can be described as: 
\begin{equation}
    \mathop{\arg\min}\limits_{\theta}\mathcal{L}'\bigl( \mathcal{F}_{\theta}(\mathbf{I}_T, \mathbf{P}_T), \mathbf{L}_T \bigr),
\end{equation}
where $\mathbf{P}_T$ denote input prompt sets comprised of randomly generated prompts extracted from annotations $\mathbf{L}_T$. $\theta$ is the trainable parameters of SAM model $\mathcal{F}$. Specifically, $\theta$ corresponds to the parameters of prompt encoder. $\mathcal{L}'$ denotes the loss function, such as cross-entropy loss, Dice loss, $\textit{etc}$.

By learning a new prompt encoder, we redefine the semantics of the down-stream segmentation tasks, shifting from the general foreground/background vision segmentation to specific medical segmentation, for instance, kidney/background, liver/background segmentation.

\subsubsection{$\text{SAM}^{\text{auto}}$}
In contrast to the \assist model, the \auto model is designed to automatically generate the annotations without user interaction. \auto aims to further boost medical annotation procedures. The architectural framework of \auto is illustrated at \cref{fig:sam_auto}. Instead of relying on manual input prompts from users, we introduce a ``prompt generator'' module, denoted as $\mathcal{G}$, which automatically generates potential prompts for the query images. This module is trained on a limited number of manually annotated slices. The image segmentation module employed in the \auto is the pre-trained \assist model. Therefore the objective of \auto module can be described as follows:
\begin{equation}
    \mathop{\arg\min}\limits_{\omega}\mathcal{L}'\bigl( \mathcal{F}(\mathbf{I}_T, \mathcal{G}_{\omega}(\mathbf{I}_T)), \mathbf{L}_T \bigr).
\end{equation}
Consequently, the primary challenge lies in developing an app)ropriate prompt generator for the SAM model. There are various approaches can be explored for generating the prompts for query image slices. In this study, we mainly investigated three approaches: 

\begin{figure}[b]
\begin{center}
   \includegraphics[width=\linewidth]{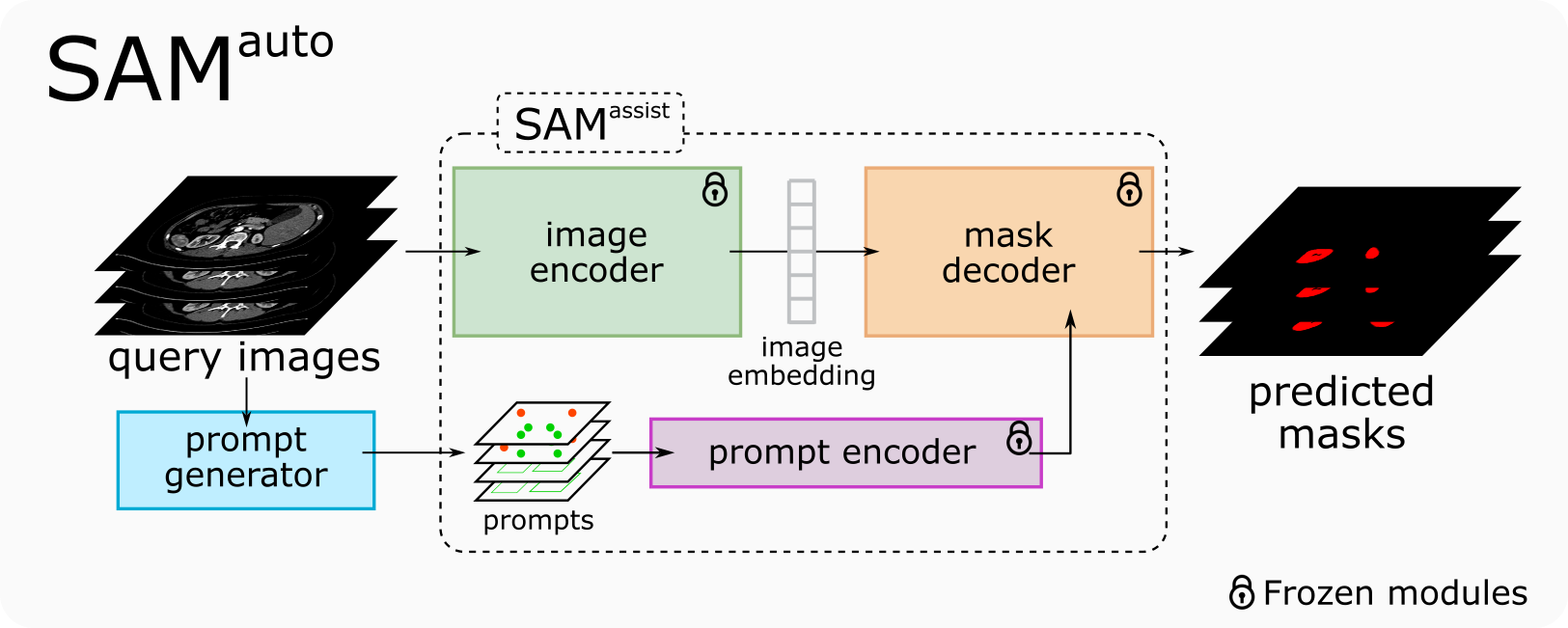}
\end{center}
   \caption{Overview of the proposed \auto model. The employed \assist model is frozen during the training. Only the ``prompt generator'' model is trainable.
   }
\label{fig:sam_auto}
\end{figure}

\vspace{-10pt}
\paragraph{Prompt propagation} Taking inspiration from numerous label propagation approaches, prompt propagation is a straightforward and intuitive approach to generate potential prompts for new images. Considering the consistency between medical image slices, prompt propagation can effectively deduce the possible prompts based on annotated seed prompts. Nevertheless, one of the key challenges in employing the prompt propagation approach is how to establish an appropriate propagation criterion. Additionally, several inherent drawbacks of prompt propagation remain to be addressed. Firstly, it cannot handle discrete objects, such as nuclei and lung nodule segmentation tasks. Secondly, propagating the background prompts is difficult due to its diverse tissues composition.

In this work, we implemented a simple prompt propagation strategy by dynamic thresholding approach. The specific algorithmic details can be found in \Cref{alg:pt_propagate}. The foreground points will be recursively propagate to the neighboring slices based on intensity consistency principles.

\begin{algorithm}[h]
    \SetAlgoLined
    \KwData{Image slice $X_t \in \mathbb{R}^{H\times W}$ and corresponding annotated mask $L_t\in \{0, 1\}^{H\times W}.$ Adjacent slices $X_{t+1}$. Pre-trained \assist model $\mathcal{F}_{assist}(\cdot)$.}
    \KwResult{New prompts $\mathbf{P}_{t+1}$ on $X_{t+1}$}
    
    \SetKwFunction{algorithmName}{PromptPropagate}
    \SetKwProg{Fn}{Function}{:}{}
    
    \Fn{\algorithmName{$X_t, L_t, X_{t+1}$}}{
        $\tilde{X} = \text{stddev}(X_t(L_t))$ \;
        $\mathbf{P}_t = \text{random\_sample}(L)$ \;
        \For{i = 1,...,n}{
            \If{$| X_{t+1}(\mathbf{P}_t^i) - X_t(\mathbf{P}_t^i) | < \lambda\tilde{X}$}{
                $\mathbf{P}_{t+1}^i = \mathbf{P}_t^i$\;
            }
            $L_{t+1} = \mathcal{F}_{assist}(X_{t+1}, \mathbf{P}_{t+1})$ \;
        }
        \Return $\mathbf{P}_{t+1}$\;
    }
    \caption{Prompt propagation}
    \label{alg:pt_propagate}
\end{algorithm}

\vspace{-10pt}
\paragraph{Prompt classification} An alternative approach to generate prompts is prompt classification, which directly categorizing prompts (preferably point prompt) into foreground and background classes. This approach leverages the inherent characteristics of annotated slices to train a straightforward classifier. By analyzing the image features extracted by SAM's well-trained image encoder, the classifier learns to differentiate between foreground and background prompts.

Compared to the prompt propagation method, classification-based method provide a more systematic and flexible way to generate prompts. By capitalizing on the wealth of information present in the annotated data, the classifier can be easily applied to other unlabeled data without seed prompts. Despite the merits, prompt classification also presents certain challenges. Designing an effective classifier that can suppress the over-fitting issue on the limited annotated data is the key point of prompt classification approaches.

\vspace{-10pt}
\paragraph{Coarse segmentation} In addition to the previously mentioned methods, we also investigated the feasibility of directly segment the object from the query image slices as part of the prompt generation process. Coarse segmentation offers another alternative approach to prompt generation by providing a coarse spatial localization of the foreground. By segmenting the objects directly, we aim to capture their boundaries and spatial extent, which can be useful in generating prompts that are more closely aligned with the target objects of interest.

\begin{figure*}[tb]
\begin{center}
   \includegraphics[width=\linewidth]{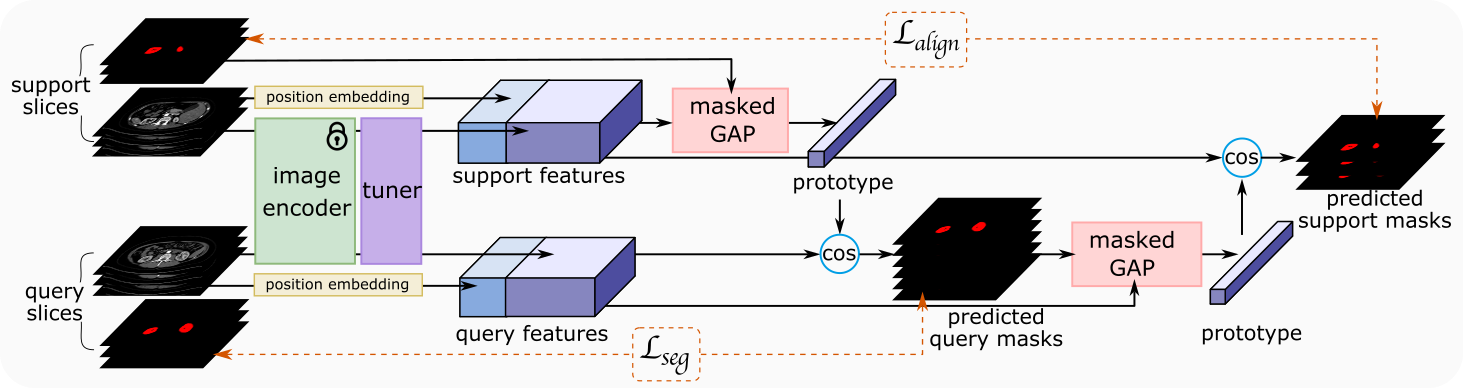}
\end{center}
   \caption{Overview of our proposed SAP-Net.
   }
\label{fig:segmentor}
\end{figure*}

In contrast to prompt propagation and prompt classification, which focus on the generating prompt itself, coarse segmentation aims to locate and extract relevant features specific to the objects of interest by explicitly delineate the foreground region. Notice that the objective of coarse segmentation is not to replace the SAM's mask decoder, it attempts to leverage the limited annotations to locate the target object and generate accurate prompts for the well-trained SAM model. 

In this work, we designed a lightweight ROI segmenter using few-shot learning technique. Compared to conventional fully convolutional networks (FCNs), such as U-Net \cite{ronneberger2015u} and deeplab, which were designed for fully supervised fine segmentation, few-shot segmentation approaches aims to address the challenges of limited labeled data. Our proposed spatial-aware prototypical net, namely SAP-Net, attempts to leverage the positional consistency between adjacent medical slices and the power of prototype learning to accurately locate the target regions. The architecture of the SAP-Net is illustrated at \cref{fig:segmentor}. We used the PANet \cite{wang2019panet} as our backbone. PANet is one variant of Prototypical Network \cite{snell2017prototypical} which introduced an auxiliary alignment loss to regularize the prototypes on both support and query sets.

In the training phase, all annotated slices $\mathcal{D}_{Train} = \{(I_T^i, L_T^i)\}_{i=1}^N$ will be divided into support set $\mathcal{S} = \{(I_s^j, L_s^j)\}_{j=1}^J$ and query set $\mathcal{Q} = \{(I_q^k, L_q^k)\}_{k=1}^K, J+K=N$. Instead of fine-tuning the SAM's pre-trained image encoder directly, we added a small trainable FCN network to tune the features from the pre-trained image encoder. Following the same architecture of PANet, we use both segmentation loss $\mathcal{L}_{seg}$ and $\mathcal{L}_{align}$ as total criterion to optimize the network. $\mathcal{L}_{seg}$ encourages high segmentation quality for the query images using the prototypes from support set. Conversely, $\mathcal{L}_{align}$ encourages the prototypes generated from the queries to align well with those of the support sets. Concretely, $\mathcal{L}_{seg}$ can be formulated as follows:

\begin{equation}
    \mathcal{L}_{seg} = - \frac{1}{K} \sum_{k\in K} d(\hat{L}_q^j, L_q^j),
\end{equation}
where $d$ denotes any distance measurement such as Dice distance, cross-entropy distance. $\hat{L}_q$ denotes predicted mask for the query image which is computed as follows:
\begin{equation}
    \hat{L}_{q;c}^{(x,y)} = \mathop{\arg\max}\limits_{c} \frac{\exp(-\alpha \cos(\mathbf{F}_q^{(x,y)}, p_c))}{\sum_{p_c \in \mathcal{P}}\exp(-\alpha\cos(\mathbf{F}_q^{(x,y)}, p_c))},
\end{equation}
$\cos$ denotes cosine distance which is used to measure the distance between prototype $p_c$ and image features $\mathbf{F}_q$. $\mathcal{P} = \{ p_c | c \in \mathcal{C}_i \}$ is prototype set contains all class prototypes. $\alpha$ is a weight factor. The prototype is computed via masked global average pooling \cite{wang2019panet}.

To fully leverage the positional consistency between adjacent slices, we presented spatial-aware functionality by introducing a position encoding layer \cite{tancik2020fourier}. Therefore, the image feature $\mathbf{F}_q$ can be obtained by:
\begin{equation}
    \mathbf{F}_q^{(x,y)} = \bigl[\mathcal{T}\Bigl(\mathcal{E}\bigl( \mathbf{I}_q^{(x,y)} \bigr) \Bigr), \gamma(x, y)\bigr],
\end{equation}
where $\mathcal{E}$ is the pre-trained image encoder of SAM, $\mathcal{T}$ is the additional FCN network to tune the image features from $\mathcal{E}$. $[\cdot,\cdot]$ denotes the concatenate operation. $\gamma(\cdot)$ is the position encoder layer which is computed as follows:
\begin{equation}
    \gamma(\mathbf{x, y}) = [\cos(2\pi\mathbf{B\cdot(x, y)}), \sin(2\pi\mathbf{B\cdot(x, y)})]^\top ,
\end{equation}
where $\mathbf{B} \in \mathbb{R}^{2\times d}$ is sampled from $\mathcal{N}(0, \sigma^2)$ distribution. $d$ is the position embedding feature dimension. Vice versa, all previous calculations are same for the $\mathcal{L}_{align}$ regards to the support set. 

Furthermore, in contrast to conventional few-shot segmentation tasks, the coarse segmentation is specifically designed to automatically generate prompt candidates for the \assist. Consequently, false positives (FPs) potentially have a more detrimental impact than false negatives (FNs). To mitigate the occurrence of false positives (FPs), we introduced a biased Dice loss that imposes a higher penalty on FPs. The loss function could be formulated as:
\begin{equation}
    \mathcal{L} = 1 - \dfrac{2\times TP}{2\times TP + \beta FP + FN},
\end{equation}
where TP (True Positives) represents the number of pixels that were correctly predicted as positive in the segmentation. FP represents the number of pixels that were incorrectly predicted as positive.
FN represents the number of pixels that were incorrectly predicted as negative. $\beta$ is the bias weight for the FP term.

\subsection{Training \& Validation}\label{train_valid}
This section provides a detailed description of the training and validation processes. The training of the \assist model involves three distinct prompts: points, boxes, and their combination. The utilization of individual prompt aims to showcase the overarching segmentation ability of \assist. Notably, the composite prompt comprising both points and boxes serves as a emulation of real-world annotation scenarios encountered in routine usage of \assist for the purpose of annotating medical images in everyday clinical practice.

\begin{figure}[b]
    \centering
    \begin{subfigure}{0.3\linewidth}
        \includegraphics[width=\textwidth]{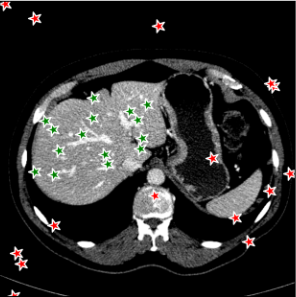}
        \caption{Uniform}
    \end{subfigure}
    \begin{subfigure}{0.3\linewidth}
       \includegraphics[width=\textwidth]{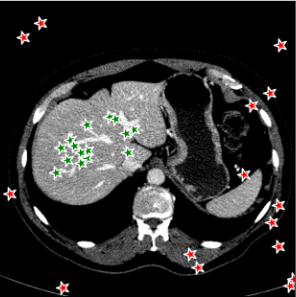}
       \caption{Center}     
    \end{subfigure}
    \begin{subfigure}{0.3\linewidth}
        \includegraphics[width=\textwidth]{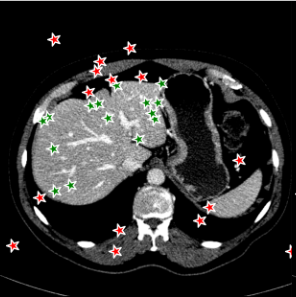}
        \caption{Boundary}
    \end{subfigure}
    \caption{Three point sampling schemes. 15 points were randomly sampled in both foreground (liver) and background regions with different sampling schemes.}
    \label{fig:pt_sampling}
\end{figure}

For both \assist and \auto experiments, the training data $\mathcal{D}_{Train} = \{(I_T^i, L_T^i)\}_{i=1}^N$ are generated with a random selection strategy, implemented to emulate user behavior in a more verisimilar manner. Notably, annotated slices are randomly chosen using a Normal distribution, characterized as $(I_T^i, L_T^i) \sim \mathcal{N}(m, s^2)$, where $m$ is the median slice within the cohort of all foreground slices, $s$ denotes the standard deviation of foreground ranges. Meanwhile, the background slices which needed for the \auto model are randomly selected within the background slices following a uniform distribution.

\vspace{-1em}
\subsubsection*{Point prompt}
Points are the most fundamental interactions in semi-automatic annotation tools, but also most unpredictable. To train a new point-based prompt tailored to a specific annotation task, a stochastic process is employed wherein $n_{pt}$ points are randomly sampled from the interval of $[1, N_{pt})$ within both the background and foreground regions of each annotated slice during every iteration of the training process. In this work, three distinct point selection schemes has also been investigated, namely `uniform', `center' and `boundary'. An example is illustrated in \cref{fig:pt_sampling}. In the case of uniform sampling, points are random selected with uniform distribution within the mask region. For center sampling, we primarily selected the points at the central location within the target region, by applying a distance transform. Conversely, the boundary sampling mainly selects the point near the boundary.

In the validation stage, we also randomly selected $n_{pt}$ points on both foreground and background to simulate user's manual click. Given these input points, the predicted mask will be compared to the ground truth for performance evaluation.

\begin{figure}[b]
  \centering
  \begin{minipage}[t]{0.45\linewidth}
    \centering
    \begin{adjustbox}{valign=t}
      \includegraphics[width=\linewidth]{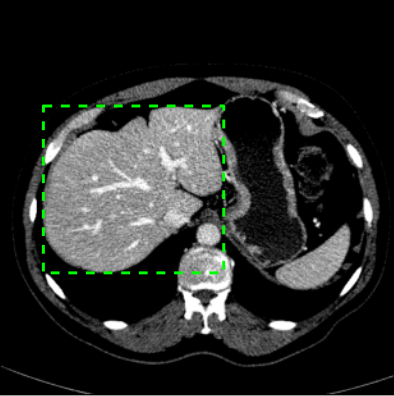}
    \end{adjustbox}
  \end{minipage}%
  \hspace{0.1em}
  \begin{minipage}[t]{0.45\linewidth}
    \centering
    \begin{adjustbox}{valign=t}
      \includegraphics[width=\linewidth]{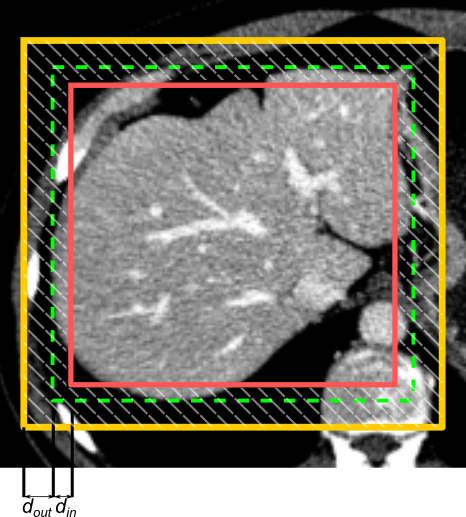}
    \end{adjustbox}
  \end{minipage}
  \caption{An illustration of box jittering during the box prompt-learning phase. The green box denotes the original bounding-box of the target tissue. The yellow and red boxes denote the outer and inner boundaries with a margin of $d_{out}$ and $d_{in}$, respectively. The jittered box will be positioned within the region delineated by the outer and inner boundaries, illustrated as the shaded region.}
  \label{fig:box_jitter}
\end{figure}

\begin{table*}[t]
    \centering
    \begin{tabular}{lp{5cm}lll}
    \toprule
    Dataset ID   & Full dataset name and citation & Modality   & Target tissue           & Case number \\
    \midrule
    Kidney-CT    & Kidney Tumor Segmentation Challenge (KiTS) \cite{HELLER2021101821}  & CT   & Kidney along with tumor & 210 \\
    Kidney-US    & CT2US for Kidney Segmentation \cite{SONG2022106706} & Ultrasound & Kidney  & 4586   \\
    Liver-CT     & The Liver Tumor Segmentation Benchmark (LiTS) \cite{BILIC2023102680} & CT  & Liver along with tumor  & 131 \\
    Lung-CT      & CT Volumes with Multiple Organ Segmentations (CT-ORG) \cite{rister2020ct} & CT & Lung &  139    \\
    Prostate-MRI & PROSTATEx Challenges \cite{prostatex}  & MRI        & Gland region, PZ region     &   346  \\
    Spleen-CT    & Medical Segmentation Decathlon \cite{antonelli2022msd}  & CT   & Spleen  & 41  \\
    Nuclear-Hist & Colonic Nuclear Instance Segmentation (Lizard) \cite{graham2021lizard} & Histopathology & Nuclear  &  238   \\
    Spine-MRI    & Spinal Cord Grey Matter Segmentation Challenge \cite{prados2017spinal} & MRI   & Spinal cord & 40 \\
    MultiOrgn-CT & Multi-Atlas Labeling Beyond the Cranial Vault (BTCV) \cite{xu2016btcv} & CT & 13 abdominal organs & 30 \\ \bottomrule
    \end{tabular}
    \caption{All dataset used in this work.}
    \label{tab:dataset}
\end{table*}

\vspace{-1em}
\subsubsection*{Box prompt}
Boxes have been reported to be the most effective prompts in SAM framework \cite{mazurowski2023segment,cheng2023sam}. Compared to points, boxes inherently provide more comprehensive information. Nevertheless, it is imperative to acknowledge that the utilization of boxes presents inherent challenges, particularly in the segmentation of non-convex objects. Additionally, boxes are notably susceptible to the box jitter \cite{cheng2023sam}. Consequently, during the phase of box prompt-learning, random box jitter in range of $[-d_{in}, d_{out}]$ has been intentionally introduced into the training stage. The introduction of box jitter aims to facilitate the training of a more robust box embedding, capable of accommodating across diverse annotation deviations. A simple illustration of the box jittering is shown in \cref{fig:box_jitter}.

In the validation stage, the box jitter has also been implemented during the inference to replicate the users' actual annotation processes, which might deviate from the ground truth.

\vspace{-1em}
\subsubsection*{Composite prompt}
Composite prompt comprising both points and boxes, is specifically designed to simulate real-world annotation scenarios. An authentic interactive annotation might involve using a box to initially segment the target, followed by using points to refine the segmentation results. \assist also support prompt learning for composite prompts. Different from previous single prompt types, points and boxes are simultaneously generated randomly during training. The sampling and jittering schemes remain consistent with those mentioned earlier.

During the validation stage, we implemented an active sampling scheme to simulate the real-world annotation process. Box prompt will be firstly placed to initially segment the target. Subsequently, foreground and background points will be continuously placed at the FN and FP regions, respectively. For each target instance, a maximum of one box and up to five points are sampled for validation.The final segmentation performance is determined based on the best Dice score achieved.

\section{Experiments and Results}
The proposed \sammed framework is composed of \assist and \auto models. Each of these models will be independently assessed to evaluate their respective performance. All experiments were conducted using PyTorch 1.13.1 in Python and executed on a single V100 GPU.

\subsection{Dataset}
Several publicly available datasets have been employed in this study for the purpose of evaluation. Comprehensive details are listed in \Cref{tab:dataset}. We collected eight public dataset encompassing various radiological modalities to effectively demonstrate the generalizability of our proposed \sammed. More detailed preprocessing of each dataset is described in Supplementary 1.

\begin{figure*}[bt]
\begin{center}
   \includegraphics[width=\linewidth]{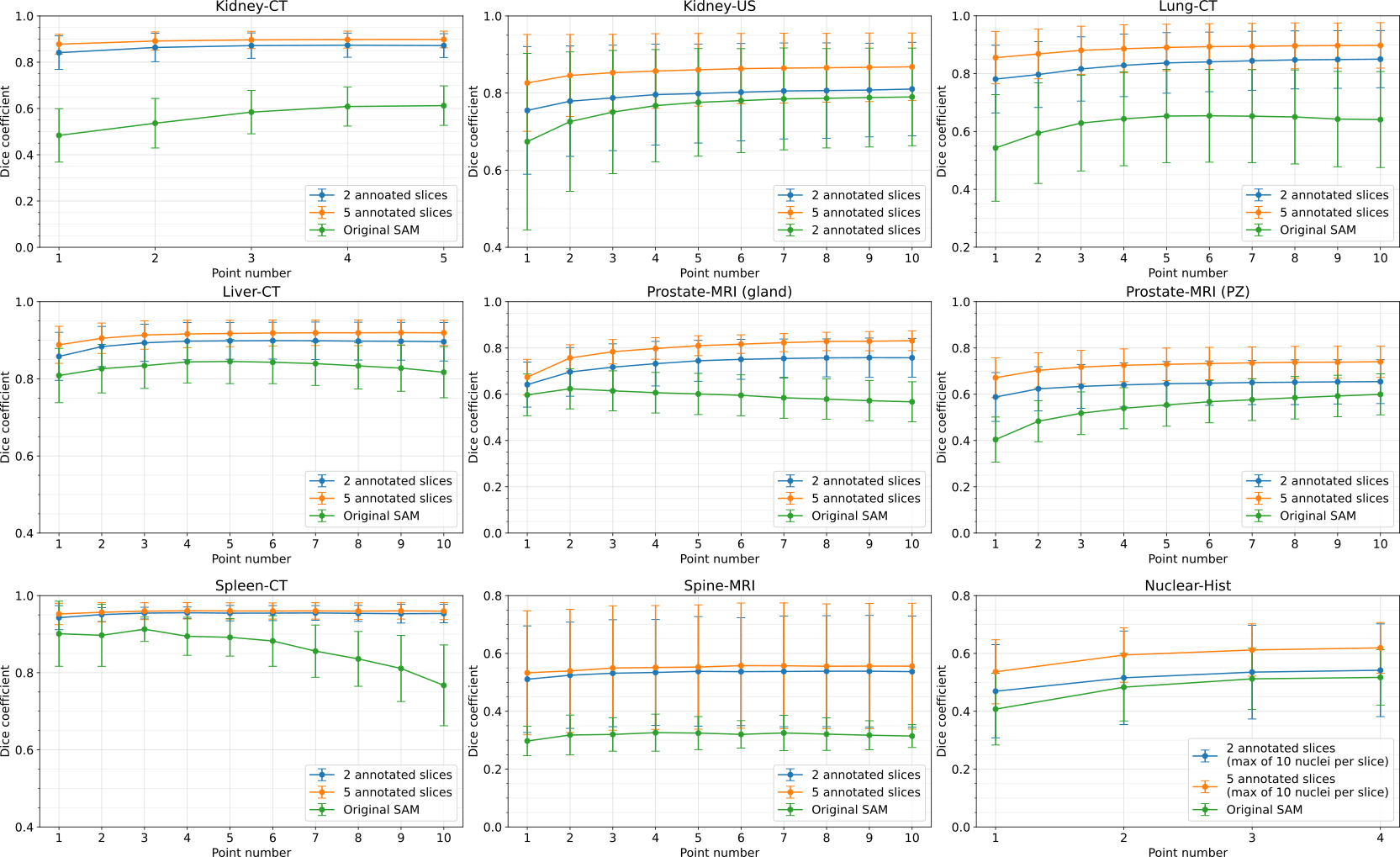}
\end{center}
   \caption{The point-based \assist experimental results were assessed by evaluating the performances on 2 and 5 annotated slices used for prompt training. The X-axis represents the number of given points per instance during the inference stage, while the Y-axis corresponds to the average Dice coefficient across all cases, accompanied by the standard deviation.}
\label{fig:assist_ret1}
\end{figure*}

\subsection{$\textbf{SAM}^{\text{assist}}$}\label{sec:experiment1}
Given that \assist is specifically designed as a semi-automatic segmentation tool, our main objective is to assess its efficacy in minimizing manual interactions. The assessment of \assist encompasses three key metrics: segmentation accuracy, interaction costs and training time.

\def\dyncaptions#1{\ifcase#1\or 2 annotated slices\or 5 annotated slices\or Original SAM\fi} 

\begin{figure*}[t]
    \centering
    \foreach \x [count=\xi] in {2_slices, 5_slices, Original_SAM}{
        \foreach \y in {1,...,6}{
            \begin{subfigure}{0.16\linewidth}
                \centering
                \begin{adjustbox}{trim=0.2cm 0.05cm 0.2cm 0.05cm, clip}
                    \includegraphics[width=\textwidth]{Figures/kits_assist_plt/\x/\y_assist_valid_pred.png}
                \end{adjustbox}
            \end{subfigure}\hspace{-1.8em}
        }\vspace{-1em}
        \caption*{\dyncaptions\xi}
    }
\caption{Segmentation results of the Kidney-CT dataset using \assist. The top and middle rows denote \assist training with 2 and 5 annotated slices, respectively. The bottom row represent the segmentation results by the original SAM. Point number denotes the input point number for each instance, \ie left kidney, right kidney and background. Points were randomly selected from the fore- and background region.}
\label{fig:assist_plot1}
\end{figure*}

For 3D dataset, the \assist model will be initially trained on several randomly selected annotated slices, and then tested on the remaining slices. For 2D dataset, such as Kidney-US and Nuclear-Hist, the \assist model will be initially trained on several randomly selected cases, and test on other cases. We randomly sampled points from both foreground and background regions to train the new prompt. Each case will be trained and tested, individually. \Cref{fig:assist_ret1} demonstrated the performance of the point-based \assist model. As the figure shows, single point on the target has shown remarkable segmentation performances using \assist even with only 2 annotated slices for the prompt training, compared to the original SAM model. Segmentation results of Kidney-CT are visualized in \Cref{fig:assist_plot1}. Segmentation results of one test slice is shown to demonstrate the significant improvement of \assist. Comparison results with other state-of-the-art segmentation results are shown in \Cref{tab:btcv} including both fully-supervised (such as nnUNet, UNetr), fine-tune based SAM segmentation method (such as MedSAM, MSA).

\begin{table*}[h]
\centering
\caption{Comparison experiments with other state-of-the-art segmentation methods over MultiOrgn-CT dataset including both fully-supervised (such as nnUNet, UNetr) and fine-tune based SAM segmentation method (such as MedSAM, MSA). Dice coefficient (\%) is used a evaluateion metric. Results of related works come from the literature.}
\resizebox{0.99\textwidth}{!}{%
\begin{tabular}{c|c|cccccccccccc|c}
\toprule
&Model           & Spleen & R.Kid & L.Kid & Gall. & Eso.  & Liver & Stom.  & Aorta & IVC  &Veins & Panc. & AG  & Ave     \\ \midrule
\multirow{7}{*}{\rotatebox{90}{Fully-supervised}} &TransUNet       & 95.2   & 92.7 & 92.9 & 66.2 & 75.7 & 96.9  & 88.9 & 92.0  & 83.3 & 79.1  & 77.5  & 63.7 & 83.8    \\
&EnsDiff         & 93.8   & 93.1 & 92.4 & 77.2 & 77.1 & 96.7  & 91.0 & 86.9  & 85.1 & 80.2  & 77.1  & 74.5 & 85.4    \\
&SegDiff         & 95.4   & 93.2 & 92.6 & 73.8 & 76.3 & 95.3  & 92.7 & 84.6  & 83.3 & 79.6  & 78.2  & 72.3 & 84.7    \\
&UNetr           & 96.8   & 92.4 & 94.1 & 75.0 & 76.6 & 97.1  & 91.3 & 89.0  & 84.7 & 78.8  & 76.7  & 74.1 & 85.6    \\
&Swin-UNetr      & 97.1   & 93.6 & 94.3 & 79.4 & 77.3 & 97.5  & 92.1 & 89.2  & 85.3 & 81.2  & 79.4  & 76.5 & 86.9    \\
&nnUNet          & 94.2   & 89.4 & 91.0 & 70.4 & 72.3 & 94.8  & 82.4 & 87.7  & 78.2 & 72.0  & 68.0  & 61.6 & 80.2    \\
&MedSegDiff      & 97.3   & 93.0 & 95.5 & 81.2 & 81.5 & 97.3  & 92.4 & 90.7  & 86.8 & 82.5  & 78.8  & 77.9 & 87.9    \\ \hline
\multirow{3}{*}{\rotatebox{90}{Original}} &SAM 1 point     & 51.8  & 68.6 & 79.1 & 54.3 & 58.4 & 46.1  & 56.2 & 61.2  & 40.2 & 55.3   & 51.1  & 35.4 & 54.8    \\
&SAM 3 points    & 62.2  & 71.0 & 81.2 & 61.4 & 60.5 & 51.3  & 67.3 & 64.5  & 48.3 & 62.8   & 56.4  & 39.5 & 63.1   \\ 
&SAM 10 points   & 78.5  & 77.4 & 86.3 & 65.8 & 67.3 & 78.5  & 76.0 & 71.2  & 56.2 & 70.3   & 65.1  & 52.8 & 70.4   \\ \hline
\multirow{2}{*}{\rotatebox{90}{\makecell{Fine-\\tune}}} &MedSAM 1 point  & 75.1  & 81.4 & 88.5 & 76.6 & 82.1 & 90.1  & 85.5 & 87.2  & 74.6 & 77.1   & 76.0  & 70.5 & 80.3   \\
&MSA 1 point     & 97.8  & 93.5 & 96.6 & 82.3 & 81.8 & 98.1  & 93.1 & 91.5  & 87.7 & 81.1   & 76.7  & 80.9 & 88.3   \\ \hline
\multirow{3}{*}{\rotatebox{90}{\makecell{Prompt-\\learning\\(2 slices)}}} &\assist 1 point  & 90.2  & 87.6 & 87.4 & 70.7 & 59.8 & 85.7  & 66.3 & 88.1  & 67.5 & 59.7  & 54.3  & 48.6 & 70.3   \\
&\assist 3 points  & 92.3  & 89.4 & 88.5 & 80.0 & 74.3 & 89.4  & 79.8 & 90.8  & 79.6 & 67.3  & 60.6  & 64.5 & 77.0   \\
&\assist 10 points & 91.7  & 89.8 & 88.4 & 81.8 & 80.0 & 89.0  & 82.9 & 91.3  & 83.8 & 70.3  & 63.8  & 58.6 & 79.2  \\
&\assist 1 box & 94.3 & 91.0 & 91.8 & 87.4 & 84.9 & 93.0 & 86.1 & 92.0 & 88.1 & 81.0 & 77.0 & 68.1 & 86.2 \\ \bottomrule
\end{tabular}}\label{tab:btcv}
\end{table*}

\begin{figure*}
    \centering
    \includegraphics[width=0.8\textwidth]{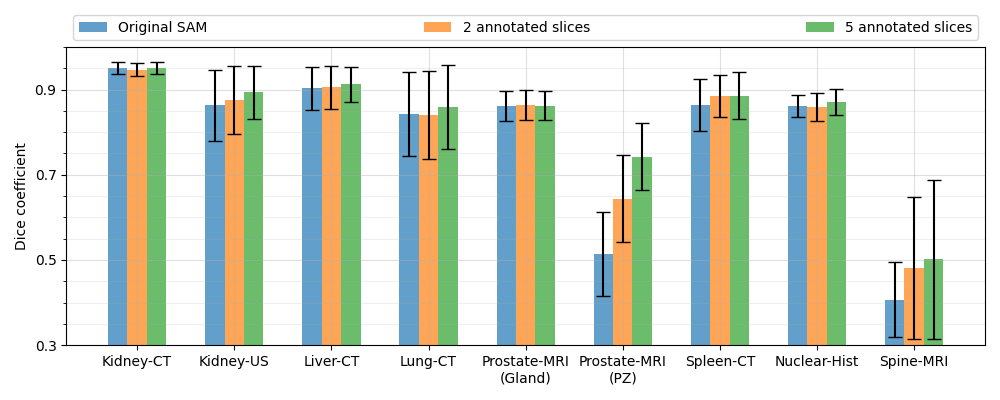}
    \caption{The box-based \assist experimental results were assessed by evaluating the performances on 2 and 5 annotated slices used for prompt training. The bar and error bar indicate the average and standard deviation of Dice coefficient.}\label{fig:assist_ret2}
\end{figure*}

The segmentation performance of the box-based \assist model is demonstrated in \Cref{fig:assist_ret2}. To simulate real manual interaction, additional box jitter was introduced to all experiments during the inference phase. Notably, the improvement of the box-based \assist model is associated with the difficulty of the task. Particularly, more substantial improvements were observed in scenarios involving challenging segmentation tasks, such as Prostate-MRI (PZ) and Spine-MRI.

The active segmentation results of the composite-prompts based \assist model are presented in \Cref{tab:composite-ret}. We selected four datasets, which retain potential for further improvement, for the purpose of evaluating the efficacy of the composite prompts. The \assist model is trained using two annotated slices for all comparison experiments trials.

\begin{table*}[h]
\centering
\caption{The average time costs of the training stage of \assist per case.}
\label{tab:assist_time}
\begin{tabularx}{\linewidth}{@{}llXXXXXXXX}
\toprule
                        & & \multicolumn{8}{c}{Time (\textit{sec.})}              \\ 
                                        &   & Kidney-CT   &Kidney-US & Liver-CT     &Lung-CT     & Prostate-MRI & Spleen-CT  &Nuclear-Hist  &Spine-MRI \\ \midrule
\multirow{2}{*}{Points} & 2 annotated slices & 11.5$\pm$4.8  & 15.0   & 10.4$\pm$5.0 &13.0$\pm$7.0  & 9.6$\pm$4.3  & 7.5$\pm$3.5 & 4.9$\pm$2.4 &6.6$\pm$2.5 \\ 
                        & 5 annotated slices & 33.4$\pm$11.1 & 27.0   & 29.5$\pm$12.2&36.4$\pm$14.0 & 22.7$\pm$9.5 & 25.2$\pm$10.7 & 18.7$\pm$8.4 &7.7$\pm$4.0 \\
\multirow{2}{*}{Boxes}  & 2 annotated slices & 10.8$\pm$5.1  & 5.0    & 7.2$\pm$3.8  &14.6$\pm$7.3   & 4.5$\pm$1.7 & 10.6$\pm$5.0 & 17.5$\pm$14.8  &4.5$\pm$2.2 \\ 
                        & 5 annotated slices & 25.4$\pm$10.2 & 18.4   & 21.5$\pm$11.5&44.0$\pm$15.5  & 14.2$\pm$5.3& 18.9$\pm$8.8 &57.5$\pm$49.0 &5.7$\pm$2.7 \\ \bottomrule
\end{tabularx}
\end{table*}

\begin{figure}
    \centering
    \includegraphics[width=0.9\linewidth]{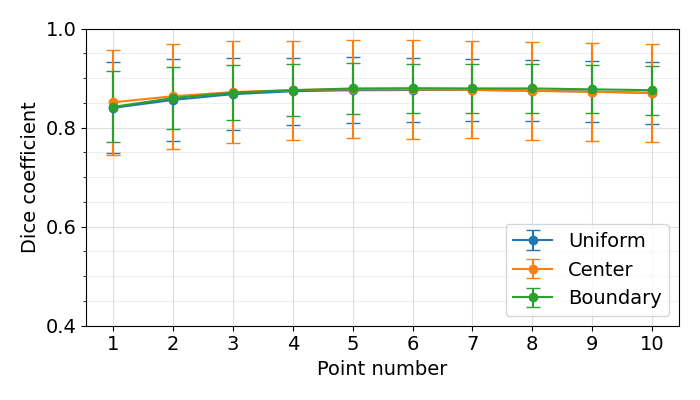}
    \caption{The experimental results regarding the influence of different point selection schemes in \assist on the Kidney-CT dataset.}
    \label{fig:point_scheme}
\end{figure}

\begin{table}[]
\caption{Segmentation results of the \assist model using composite prompts. The evaluation is conducted on four selected datasets, which may exists potential space for improvement. The \assist is trained using two annotated slices in all experimental scenarios.}
\label{tab:composite-ret}
\begin{tabularx}{\linewidth}{p{6em}ccc}
\toprule
                     & \makecell[tc]{Points\\(5 points)} & Boxes & Composite \\ \midrule
Prostate-MRI (Gland) & 74.5$\pm$8.9      & 86.3$\pm$3.5   & 88.0$\pm$3.3    \\
Prostate-MRI (PZ)    & 64.5$\pm$9.6      & 64.4$\pm$10.2  & 71.6$\pm$8.6    \\
Spine-MRI            & 53.8$\pm$18.9     & 48.1$\pm$16.6  & 58.1$\pm$3.9    \\
Lung-CT              & 83.7$\pm$10.5     & 84.1$\pm$10.3  & 91.8$\pm$6.7    \\ \bottomrule
\end{tabularx}
\end{table}

As previously mentioned, prompt-learning approach requires less resources compared to fine-tuning method. Not only data requirements but also the time costs, \Cref{tab:assist_time} provides an overview of the time costs associated with the four segmentation tasks. The table demonstrates that the time cost of point-based \assist is approximately 10 seconds and 25 seconds for the 2 and 5 annotated slices, respectively. Considering that the annotation procedure involves human-in-the-loop, the training time cost at the level of seconds is unlikely to have a substantial impact on the user experience.

The segmentation results of different point selection schemes for the Kidney-CT dataset is demonstrated in \Cref{fig:point_scheme}. Uniform random selection scheme shows slightly better variation compared to other scheme but there is no significant differences between each scheme entirely.

\subsection{$\textbf{SAM}^{\text{auto}}$}
In this work, we mainly investigated `prompt propagation' and `coarse segmentation' approaches for the explicit prompt generation. 

\begin{figure}[bt]
    \centering
    \includegraphics[width=0.8\linewidth]{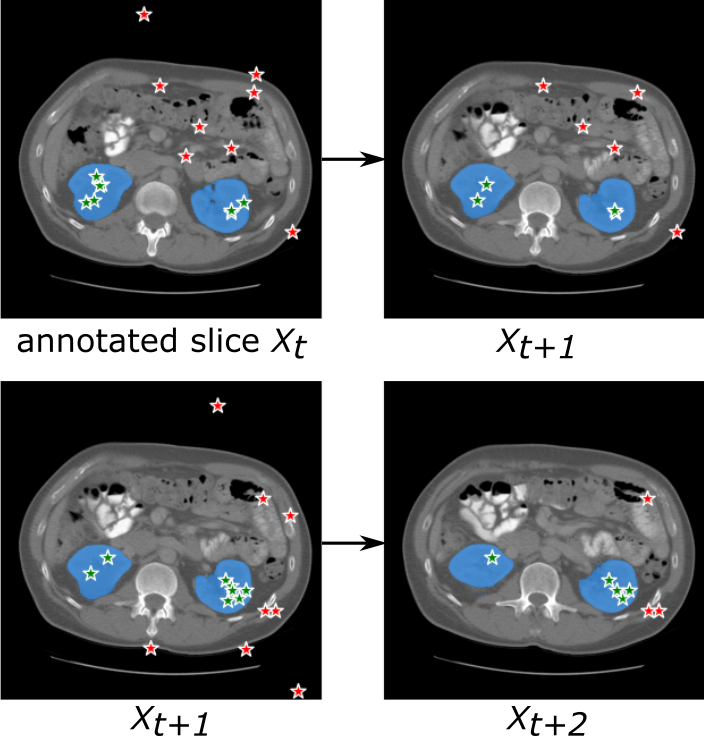}
    \caption{Illustration of point propagation procedure. The random sampled input points shown in the left column will be propagated to the next slice by a certain criteria, the propagated points are illustrated at right column.}
    \label{fig:propagate_procedure}
\end{figure}

As for the `prompt propagation' strategy described in \Cref{alg:pt_propagate}, the points that are randomly selected from the annotated slices will undergo propagation to the neighboring slices utilizing a straightforward thresholding rule. This iterative procedure will terminate when there are no points satisfying the specified criterion. To improve the robustness, the propagation is individually applied to each annotated slice. Consequently, in the case of two annotated slices, the propagation is executed twice, and the resulting segmentation are then ensemble to obtain the final segmentation. One example of point propagation procedure is shown in \Cref{fig:propagate_procedure}. 

\begin{figure*}
    \centering
    \includegraphics[width=0.8\linewidth]{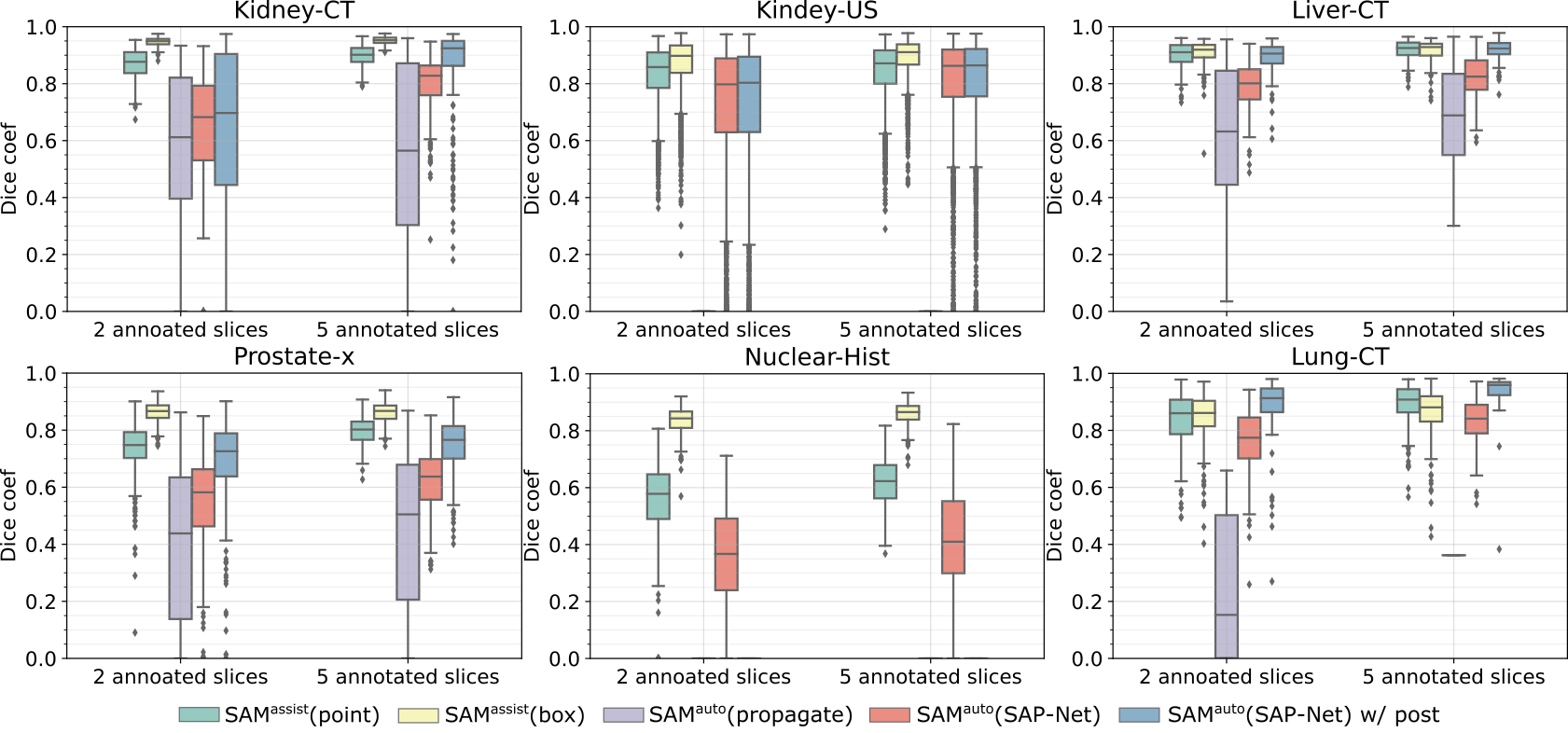}
    \caption{Comparison of different segmentation models. Green and yellow bars denote the \assist model with points and boxes prompts, respectively. Purple bar denotes the prompt propagation method. Red bar represents the \auto model, and blue bar demonstrated results with post-processing. Notice that 2D dataset like Kidney-US and Nuclear-Hist cannot use propagation method to generated the candidate prompts.}
    \label{fig:auto_ret1}
\end{figure*}

\begin{figure*}
    \centering
    \includegraphics[width=0.9\linewidth]{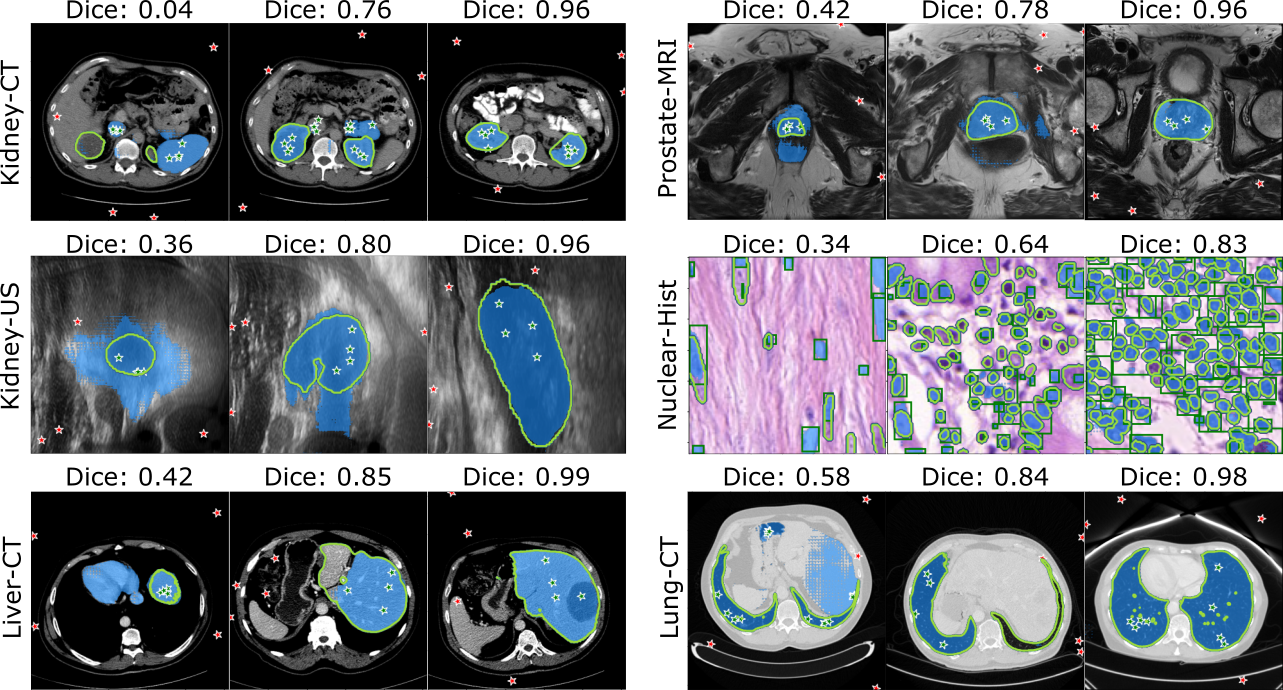}
    \caption{Visualization of \auto segmentation results. Blue masks denote the segmentation results and green contours present the ground-truth. For each dataset, it displays three examples from left to right, illustrating segmentation performances ranging from poor to excellent.}
    \label{fig:auto_vis}
\end{figure*}

As for the `coarse segmentation' strategy, annotated slices will be divided into support and query sets in the training stage. In the inference stage, all annotated slices will be used as support set. Additionally, during the inference phase, we employ a post-processing step by extracting the top-K components from the coarse segmentation results to further mitigate FPs. The segmentation results of the \auto are depicted in \Cref{fig:auto_ret1}, while the visualization of these segmentation results can be observed in \Cref{fig:auto_vis}. For each dataset, it presents three examples from left to right, progressing from left to right, illustrating segmentation performances ranging from poor to excellent. Noticed that all dataset used points as input prompts except for the Nuclear-Hist dataset, which used boxes as prompts.

Furthermore, considering that positional consistency can only be used between adjacent slices in 3D data, we specifically applied SAP-Net to the 3D dataset. In the case of 2D dataset, the baseline PANet \cite{wang2019panet} was employed. Ablation study was conducted to demonstrate the effectiveness of our SAP-Net model in the proposed \auto. The results are list in \Cref{tab:abilation}.

\begin{table}
\centering
\caption{Ablation study of our proposed SAP-Net in \auto model. Comparison experiments are conducted on Kidney-CT dataset with five annotated slices for training. ``PE" denote the introduction of position-encoding module, ``biasDice" denotes the introduced biased Dice loss, ``Post" denotes the usage of post-processing which extracts the top-K largest components.}
\label{tab:abilation}
\newcommand{\cross}{{\XSolidBrush}}
\newcommand{\checkm}{{\Checkmark}}
\newcolumntype{C}[1]{>{\centering\arraybackslash}p{#1}}
\begin{tabularx}{\linewidth}{C{3em}C{1em}C{3em}C{3em}X}
\toprule
                    & PE     & biasDice & Post     & Dice (\%) \\  \midrule
\multicolumn{1}{c}{Vanilla PANet} & \cross & \cross  & \cross  & 73.2 (11.1) \\ \hline
\multirow{4}{*}{SAP-Net} & \cross      & \cross     & \cross   &  73.9 (7.7)  \\
                         & \checkm   & \cross     & \cross     & 76.0 (7.3)   \\
                         & \checkm   & \checkm   & \cross      & 79.3 (10.8)  \\
                         & \checkm   & \checkm   & \checkm     & 84.0 (20.0)  \\ \bottomrule
\end{tabularx}
\end{table}

\section{Discussions}
In this study, we presented \sammed as an enhanced framework for medical image annotation, leveraging the capabilities of a large-scale vision model (SAM). The performance evaluation consists of two parts.

Firstly, we evaluated the \assist model to show the generalization ability of SAM model for downstream medical segmentation task by using prompt-learning approach. The results presented in \Cref{fig:assist_ret1} indicate a substantial improvement achieved through prompt-learning. Notably, the \assist model achieves an acceptable level of accuracy with only approximately five input points, using only two annotated slices for the prompt-learning. The experimental results have validated that the potential of prompt-learning with large vision model to enhance downstream tasks such as medical segmentation. In \Cref{tab:btcv}, we present the performance of \assist model trained on 2 slices. It is inspiring to observe that point-based \assist approach achieved performance that is quite comparable to fine-tuning approaches such as MedSAM and MSA. It's worth noting that fine-tuned approaches need fully labeled data to fine-tune the SAM model, while \assist requires a minimum of just 2 annotated slices. Furthermore, box-based \assist even demonstrates compatible performances with various the-state-of-the-art fully-supervised methods.

However, when employing the box prompt, as depicted in \Cref{fig:assist_ret2}, it is interesting to note that the segmentation performance exhibited variation. The original box-based SAM has already demonstrated its remarkable segmentation performances on medical images compared to point prompt, aligning with previous studies \cite{mazurowski2023segment, cheng2023sam}, such as Kidney-CT, Liver-CT and Prostate-MRI (gland). On these datasets, \assist only brings marginal improvement. In contrast, on more challenging tasks such as Prostate-MRI (PZ) and Spine-MRI, the improvement is quite substantial. We hypothesize that this discrepancy arises from the fact that on simpler tasks, image representation is more likely to be the limitation rather than prompt representation. For the most segmentation tasks, box prompt appears to be a more effective prompt compared to the point prompt. Consequently, when transitioning to the downstream tasks, point prompt is more like to be the limiting factor than the image representation which can explain the significant improvement of point-based \assist. In this situation, fine-tuning the large vision model \cite{li2023polyp,ma2023segment,cheng2023sammed2d,wu2023msa} could be one potential solution.

Furthermore, the time costs presented in \Cref{tab:assist_time} showcased the feasibility of incorporating the \assist model into a human-in-the-loop annotation procedure. In comparison to the fine-tuning approach \cite{li2023polyp,ma2023segment,cheng2023sammed2d,wu2023msa}, our proposed \assist leveraging prompt-learning approach proves to be more suitable for such annotation tasks, requiring extremely low annotation and computation resources.

\begin{figure}
    \centering
    \includegraphics[width=0.8\linewidth]{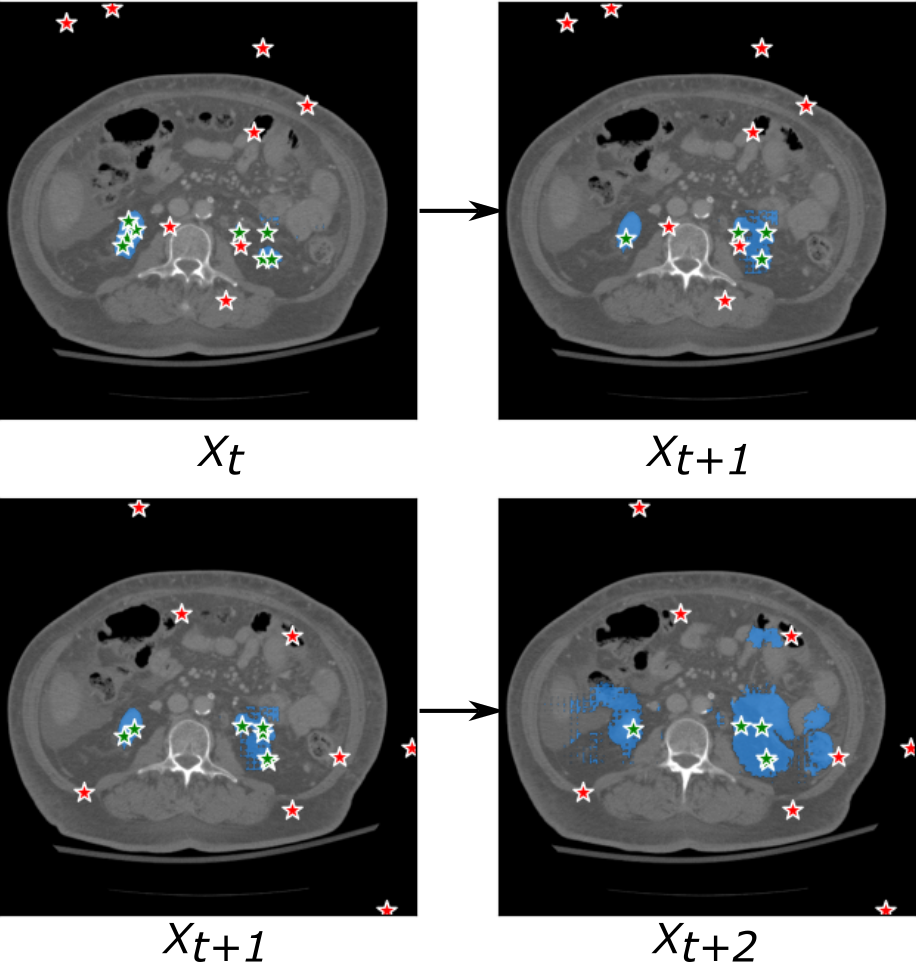}
    \caption{An example of the prompt propagation strategy's tendency of over-segmentation.}
    \label{fig:propagate_error}
\end{figure}

Secondly, we investigated the effectiveness of the proposed \auto model in further accelerating the annotation process. In this study, we presented a straightforward `prompt propagation' strategy that utilizes a simple thresholding criterion. However, the results shown in \Cref{fig:auto_ret1} indicate that this strategy did not yield satisfactory performance. We identified that the success of the propagation strategy heavily relies on the selection of appropriate stop criteria. As illustrated in \Cref{fig:propagate_error}, an inappropriate criterion led to severe over-segmentation in tissues with complex anatomical structure, such as kidney. The propagation strategy exhibited notably improved performance in liver segmentation task due to its simple anatomical structure.

In comparison, our proposed \auto with SAP-Net backbone demonstrated superior performance. Trained with only five annotated slices, \auto achieved compatible performance with \assist, which is semi-automatic approach. For instance, average Dice coefficients of 0.84, 0.92 and 0.91 were achieved on Kidney-CT and Liver-CT and Lung-CT datasets, respectively. However, there remains space for improvement on challenging Nuclear-Hist dataset. 

\begin{figure}[b]
    \centering
    \includegraphics[width=\linewidth]{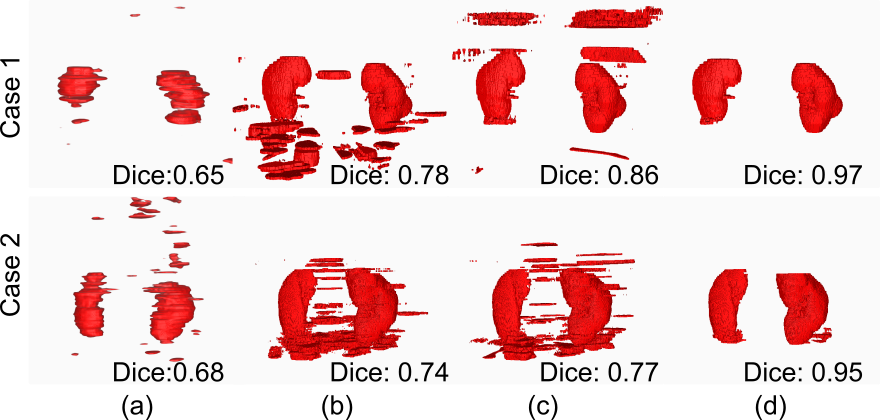}
    \caption{Ablation results of \auto. (a) represents the results of vanilla PANet. (b) denotes the reults of PANet with SAM's image encoder as backbone. (c) represnets the results of SAP-Net. (d) shows results of SAP-Net using biased Dice loss along with post-processing.}
    \label{fig:fs_slices}
\end{figure}

Although we introduced biased Dice loss to suppress the FPs, the FPs still remain as the biggest issue of the few-shot based approach. Consequently, by implementing a simple post-processing, the final segmentation performance of large solid organs exhibits a remarkable improvement, as shown in \Cref{fig:auto_ret1}. However, this simple post-processing can be only applied to the targets with a fixed number. \Cref{tab:abilation} demonstrated the effectiveness of our introduced features, successfully improve the baseline (vanilla PANet) from 73.2\% to 84.0\%. Several segmentation results of Kidney-CT are visualized in \Cref{fig:fs_slices}.

Lastly, several limitation of \auto need to be discussed. First, although \auto achieves relatively high segmentation accuracy, it may not fully satisfy annotation requirements in real-world scenarios. Annotators may still need to manually verify all the slices to confirm the segmentation. It will reduce the efficiency of the \auto. Second, as previously mentioned, the SAP-Net leveraged the spatial consistency between medical slices by incorporating position embedding. Therefore, SAP-Net cannot be applied to slices from different cases, such as Nuclear-Hist dataset which used the baseline model (the vanilla PANet).  As part of future work, further efforts are required to explore more accurate prompt generation methods.

Overall, the findings of this study demonstrate the efficacy of \sammed as an enhanced framework for medical image annotation. We think generalizability is one fundamental characteristic of a medical annotation tool. More and more AI-assisted medical annotation tools rely on pretrained model, which will heavily diminish its ability when meet unseen modality. In this work, the prompt-learning approach used in the \assist model shows promising generalization abilities, while the SAP-Net model significantly improves the segmentation performance with a minimal number of annotated slices. These results highlight the potential of leveraging large-scale vision models in the field of medical image annotation.

\section{Conclusions}
In this study, we have introduced a novel framework, named \sammed, for addressing the medical image annotation task. Our approach capitalizes on the impressive generalization and zero-shot capabilities of the large vision model, SAM. Firstly, we employed the prompt-learning technique to train the \assist model, which serves to adapt the downstream medical segmentation task to align with the capabilities of the original SAM. The primary objective of \assist is to provide a more accurate semi-automatic segmentation tool that requires minimal user interaction. Additionally, we introduced \auto to further expedite the annotation process by automatically generating the input prompt. We investigate different generation strategies and proposed the SAP-Net, which achieves satisfactory segmentation accuracy.

Nevertheless, it is important to note that this work represents only an initial step towards our overall objectives. Further extensive experiments are required to validate the effectiveness of the \sammed framework and assess its performance in real-world medical annotation scenarios.

{
\small
\bibliographystyle{unsrt}
\bibliography{egbib.bib}

\begin{thebibliography}{10}

\bibitem{kirillov2023segany}
Alexander Kirillov, Eric Mintun, Nikhila Ravi, Hanzi Mao, Chloe Rolland, Laura
  Gustafson, Tete Xiao, Spencer Whitehead, Alexander~C. Berg, Wan-Yen Lo, Piotr
  Doll{\'a}r, and Ross Girshick.
\newblock Segment anything.
\newblock {\em arXiv:2304.02643}, 2023.

\bibitem{py06nimg}
Paul~A. Yushkevich, Joseph Piven, Heather Cody~Hazlett, Rachel Gimpel~Smith,
  Sean Ho, James~C. Gee, and Guido Gerig.
\newblock User-guided {3D} active contour segmentation of anatomical
  structures: Significantly improved efficiency and reliability.
\newblock {\em Neuroimage}, 31(3):1116--1128, 2006.

\bibitem{fedorov20123dslicer}
Andriy Fedorov, Reinhard Beichel, Jayashree Kalpathy-Cramer, Julien Finet,
  Jean-Christophe Fillion-Robin, Sonia Pujol, Christian Bauer, Dominique
  Jennings, Fiona Fennessy, Milan Sonka, et~al.
\newblock {3D} slicer as an image computing platform for the quantitative
  imaging network.
\newblock {\em Magnetic resonance imaging}, 30(9):1323--1341, 2012.

\bibitem{sofiiuk2022ritm}
Konstantin Sofiiuk, Ilya~A Petrov, and Anton Konushin.
\newblock Reviving iterative training with mask guidance for interactive
  segmentation.
\newblock In {\em 2022 IEEE International Conference on Image Processing
  (ICIP)}, pages 3141--3145. IEEE, 2022.

\bibitem{liu2022simpleclick}
Qin Liu, Zhenlin Xu, Gedas Bertasius, and Marc Niethammer.
\newblock Simpleclick: Interactive image segmentation with simple vision
  transformers.
\newblock {\em arXiv preprint arXiv:2210.11006}, 2022.

\bibitem{LUO2021102102}
Xiangde Luo, Guotai Wang, Tao Song, Jingyang Zhang, Michael Aertsen, Jan
  Deprest, Sebastien Ourselin, Tom Vercauteren, and Shaoting Zhang.
\newblock Mideepseg: Minimally interactive segmentation of unseen objects from
  medical images using deep learning.
\newblock {\em Medical Image Analysis}, 72:102102, 2021.

\bibitem{tancik2020fourier}
Matthew Tancik, Pratul Srinivasan, Ben Mildenhall, Sara Fridovich-Keil, Nithin
  Raghavan, Utkarsh Singhal, Ravi Ramamoorthi, Jonathan Barron, and Ren Ng.
\newblock Fourier features let networks learn high frequency functions in low
  dimensional domains.
\newblock {\em Advances in Neural Information Processing Systems},
  33:7537--7547, 2020.

\bibitem{radford2021clip}
Alec Radford, Jong~Wook Kim, Chris Hallacy, Aditya Ramesh, Gabriel Goh,
  Sandhini Agarwal, Girish Sastry, Amanda Askell, Pamela Mishkin, Jack Clark,
  et~al.
\newblock Learning transferable visual models from natural language
  supervision.
\newblock In {\em International conference on machine learning}, pages
  8748--8763. PMLR, 2021.

\bibitem{deng2023segment}
Ruining Deng, Can Cui, Quan Liu, Tianyuan Yao, Lucas~W Remedios, Shunxing Bao,
  Bennett~A Landman, Lee~E Wheless, Lori~A Coburn, Keith~T Wilson, et~al.
\newblock Segment anything model (sam) for digital pathology: Assess zero-shot
  segmentation on whole slide imaging.
\newblock {\em arXiv preprint arXiv:2304.04155}, 2023.

\bibitem{hu2023sam}
Chuanfei Hu and Xinde Li.
\newblock When sam meets medical images: An investigation of segment anything
  model (sam) on multi-phase liver tumor segmentation.
\newblock {\em arXiv preprint arXiv:2304.08506}, 2023.

\bibitem{wald2023sam}
Tassilo Wald, Saikat Roy, Gregor Koehler, Nico Disch, Maximilian~Rouven Rokuss,
  Julius Holzschuh, David Zimmerer, and Klaus Maier-Hein.
\newblock Sam. md: Zero-shot medical image segmentation capabilities of the
  segment anything model.
\newblock In {\em Medical Imaging with Deep Learning, short paper track}, 2023.

\bibitem{putz2023segment}
Florian Putz, Johanna Grigo, Thomas Weissmann, Philipp Schubert, Daniel
  Hoefler, Ahmed Gomaa, Hassen~Ben Tkhayat, Amr Hagag, Sebastian Lettmaier,
  Benjamin Frey, et~al.
\newblock The segment anything foundation model achieves favorable brain tumor
  autosegmentation accuracy on {MRI} to support radiotherapy treatment
  planning.
\newblock {\em arXiv preprint arXiv:2304.07875}, 2023.

\bibitem{zhou2023polyp}
Tao Zhou, Yizhe Zhang, Yi~Zhou, Ye~Wu, and Chen Gong.
\newblock Can sam segment polyps?
\newblock {\em arXiv preprint arXiv:2304.07583}, 2023.

\bibitem{li2023polyp}
Yuheng Li, Mingzhe Hu, and Xiaofeng Yang.
\newblock Polyp-sam: Transfer sam for polyp segmentation.
\newblock {\em arXiv preprint arXiv:2305.00293}, 2023.

\bibitem{qiu2023learnable}
Zhongxi Qiu, Yan Hu, Heng Li, and Jiang Liu.
\newblock Learnable ophthalmology sam.
\newblock {\em arXiv preprint arXiv:2304.13425}, 2023.

\bibitem{mazurowski2023segment}
Maciej~A. Mazurowski, Haoyu Dong, Hanxue Gu, Jichen Yang, Nicholas Konz, and
  Yixin Zhang.
\newblock Segment anything model for medical image analysis: an experimental
  study, 2023.

\bibitem{cheng2023sam}
Dongjie Cheng, Ziyuan Qin, Zekun Jiang, Shaoting Zhang, Qicheng Lao, and Kang
  Li.
\newblock Sam on medical images: A comprehensive study on three prompt modes,
  2023.

\bibitem{ma2023segment}
Jun Ma and Bo~Wang.
\newblock Segment anything in medical images, 2023.

\bibitem{make3020026}
Holger~R. Roth, Dong Yang, Ziyue Xu, Xiaosong Wang, and Daguang Xu.
\newblock Going to extremes: Weakly supervised medical image segmentation.
\newblock {\em Machine Learning and Knowledge Extraction}, 3(2):507--524, 2021.

\bibitem{GUHAROY2020101587}
Abhijit {Guha Roy}, Shayan Siddiqui, Sebastian Pölsterl, Nassir Navab, and
  Christian Wachinger.
\newblock ‘{S}queeze \& excite’ guided few-shot segmentation of volumetric
  images.
\newblock {\em Medical Image Analysis}, 59:101587, 2020.

\bibitem{cai2023orthogonal}
Heng Cai, Shumeng Li, Lei Qi, Qian Yu, Yinghuan Shi, and Yang Gao.
\newblock Orthogonal annotation benefits barely-supervised medical image
  segmentation.
\newblock In {\em Proceedings of the IEEE/CVF Conference on Computer Vision and
  Pattern Recognition}, pages 3302--3311, 2023.

\bibitem{monailabel}
Andres Diaz-Pinto, Pritesh Mehta, Sachidanand Alle, Muhammad Asad, Richard
  Brown, Vishwesh Nath, Alvin Ihsani, Michela Antonelli, Daniel Palkovics,
  Csaba Pinter, Ron Alkalay, Steve Pieper, Holger~R. Roth, Daguang Xu, Prerna
  Dogra, Tom Vercauteren, Andrew Feng, Abood Quraini, Sebastien Ourselin, and
  M.~Jorge Cardoso.
\newblock Deepedit: Deep editable learning for interactive segmentation of 3d
  medical images.
\newblock In Hien~V. Nguyen, Sharon~X. Huang, and Yuan Xue, editors, {\em Data
  Augmentation, Labelling, and Imperfections}, pages 11--21, Cham, 2022.
  Springer Nature Switzerland.

\bibitem{qu2023annotating}
Chongyu Qu, Tiezheng Zhang, Hualin Qiao, Jie Liu, Yucheng Tang, Alan Yuille,
  and Zongwei Zhou.
\newblock Annotating 8,000 abdominal ct volumes for multi-organ segmentation in
  three weeks, 2023.

\bibitem{wu2023msa}
Junde Wu, Rao Fu, Huihui Fang, Yuanpei Liu, Zhaowei Wang, Yanwu Xu, Yueming
  Jin, and Tal Arbel.
\newblock Medical sam adapter: Adapting segment anything model for medical
  image segmentation.
\newblock {\em arXiv preprint arXiv:2304.12620}, 2023.

\bibitem{cheng2023sammed2d}
Junlong Cheng, Jin Ye, Zhongying Deng, Jianpin Chen, Tianbin Li, Haoyu Wang,
  Yanzhou Su, Ziyan Huang, Jilong Chen, Lei Jiang, Hui Sun, Junjun He, Shaoting
  Zhang, Min Zhu, and Yu~Qiao.
\newblock Sam-med2d, 2023.

\bibitem{ronneberger2015u}
Olaf Ronneberger, Philipp Fischer, and Thomas Brox.
\newblock U-net: Convolutional networks for biomedical image segmentation.
\newblock In {\em Medical Image Computing and Computer-Assisted
  Intervention--MICCAI 2015: 18th International Conference, Munich, Germany,
  October 5-9, 2015, Proceedings, Part III 18}, pages 234--241. Springer, 2015.

\bibitem{wang2019panet}
Kaixin Wang, Jun~Hao Liew, Yingtian Zou, Daquan Zhou, and Jiashi Feng.
\newblock {PANet}: Few-shot image semantic segmentation with prototype
  alignment.
\newblock In {\em proceedings of the IEEE/CVF international conference on
  computer vision}, pages 9197--9206, 2019.

\bibitem{snell2017prototypical}
Jake Snell, Kevin Swersky, and Richard Zemel.
\newblock Prototypical networks for few-shot learning.
\newblock {\em Advances in neural information processing systems}, 30, 2017.

\bibitem{HELLER2021101821}
Nicholas Heller, Fabian Isensee, Klaus~H. Maier-Hein, Xiaoshuai Hou, Chunmei
  Xie, Fengyi Li, Yang Nan, Guangrui Mu, Zhiyong Lin, Miofei Han, Guang Yao,
  Yaozong Gao, Yao Zhang, Yixin Wang, Feng Hou, Jiawei Yang, Guangwei Xiong,
  Jiang Tian, Cheng Zhong, Jun Ma, Jack Rickman, Joshua Dean, Bethany Stai,
  Resha Tejpaul, Makinna Oestreich, Paul Blake, Heather Kaluzniak, Shaneabbas
  Raza, Joel Rosenberg, Keenan Moore, Edward Walczak, Zachary Rengel, Zach
  Edgerton, Ranveer Vasdev, Matthew Peterson, Sean McSweeney, Sarah Peterson,
  Arveen Kalapara, Niranjan Sathianathen, Nikolaos Papanikolopoulos, and
  Christopher Weight.
\newblock The state of the art in kidney and kidney tumor segmentation in
  contrast-enhanced ct imaging: Results of the kits19 challenge.
\newblock {\em Medical Image Analysis}, 67:101821, 2021.

\bibitem{SONG2022106706}
Yuxin Song, Jing Zheng, Long Lei, Zhipeng Ni, Baoliang Zhao, and Ying Hu.
\newblock {CT2US}: Cross-modal transfer learning for kidney segmentation in
  ultrasound images with synthesized data.
\newblock {\em Ultrasonics}, 122:106706, 2022.

\bibitem{BILIC2023102680}
Patrick Bilic, Patrick Christ, et~al.
\newblock The liver tumor segmentation benchmark (lits).
\newblock {\em Medical Image Analysis}, 84:102680, 2023.

\bibitem{rister2020ct}
Blaine Rister, Darvin Yi, Kaushik Shivakumar, Tomomi Nobashi, and Daniel~L
  Rubin.
\newblock {CT-ORG}, a new dataset for multiple organ segmentation in computed
  tomography.
\newblock {\em Scientific Data}, 7(1):381, 2020.

\bibitem{prostatex}
Samuel~G. Armato, Henkjan Huisman, Karen Drukker, Lubomir Hadjiiski, Justin~S.
  Kirby, Nicholas Petrick, George Redmond, Maryellen~L. Giger, Kenny Cha, Artem
  Mamonov, Jayashree Kalpathy-Cramer, and Keyvan Farahani.
\newblock {PROSTATEx Challenges for computerized classification of prostate
  lesions from multiparametric magnetic resonance images}.
\newblock {\em Journal of Medical Imaging}, 5(4):044501, 2018.

\bibitem{antonelli2022msd}
Michela Antonelli, Annika Reinke, Spyridon Bakas, Keyvan Farahani, Annette
  Kopp-Schneider, Bennett~A Landman, Geert Litjens, Bjoern Menze, Olaf
  Ronneberger, Ronald~M Summers, et~al.
\newblock The medical segmentation decathlon.
\newblock {\em Nature communications}, 13(1):4128, 2022.

\bibitem{graham2021lizard}
Simon Graham, Mostafa Jahanifar, Ayesha Azam, Mohammed Nimir, Yee-Wah Tsang,
  Katherine Dodd, Emily Hero, Harvir Sahota, Atisha Tank, Ksenija Benes, et~al.
\newblock Lizard: a large-scale dataset for colonic nuclear instance
  segmentation and classification.
\newblock In {\em Proceedings of the IEEE/CVF International Conference on
  Computer Vision}, pages 684--693, 2021.

\bibitem{prados2017spinal}
Ferran Prados, John Ashburner, Claudia Blaiotta, Tom Brosch, Julio
  Carballido-Gamio, Manuel~Jorge Cardoso, Benjamin~N Conrad, Esha Datta,
  Gergely D{\'a}vid, Benjamin De~Leener, et~al.
\newblock Spinal cord grey matter segmentation challenge.
\newblock {\em Neuroimage}, 152:312--329, 2017.

\bibitem{xu2016btcv}
Zhoubing Xu, Christopher~P Lee, Mattias~P Heinrich, Marc Modat, Daniel
  Rueckert, Sebastien Ourselin, Richard~G Abramson, and Bennett~A Landman.
\newblock Evaluation of six registration methods for the human abdomen on
  clinically acquired ct.
\newblock {\em IEEE Transactions on Biomedical Engineering}, 63(8):1563--1572,
  2016.

\end{thebibliography}
}

\end{document}